\documentclass{article}
\usepackage{hyperref}

\usepackage[utf8]{inputenc}

\title{SmartAPS: Tool-augmented LLMs for Operations Management \\ 
{\small AAAI25 - Combining AI and OR/MS for Better Trustworthy Decision Making}}
\date{}

\usepackage{natbib}
\usepackage{graphicx}
\usepackage{color}
\usepackage{geometry}

\newgeometry{vmargin={25mm}, hmargin={40mm,40mm}}   

\begin{document}

\maketitle

\author{
  \begin{center}
  Timothy T. Yu\textsuperscript{1},
  Mahdi Mostajabdaveh\textsuperscript{1},
  Serge J. Byusa\textsuperscript{1},
  Rindra Ramamonjison\textsuperscript{1},\\
  Giuseppe Carenini\textsuperscript{1,2},
  Kun Mao\textsuperscript{3},
  Zirui Zhou\textsuperscript{1},
  Yong Zhang\textsuperscript{1} \\[0.5em]
  \textsuperscript{1}Huawei Technologies Canada \\
  \textsuperscript{2}University of British Columbia \\
  \textsuperscript{3}Huawei Cloud Computing Technologies \\[0.5em]
  \texttt{\{timothyt.yu, mahdi.mostajabdaveh1, jabo.byusa, rindranirina.ramamonjison, zirui.zhou, yong.zhang3\}@huawei.com} \\
  \texttt{giuseppe.carenini@huawei.com}, \texttt{maokun@huawei.com} \\[3em]
  \end{center}
}

\begin{abstract}
Artificial intelligence is well-positioned to support every stage of the entire OR project life cycle. Specific to this paper, recent advancements in natural language processing has presented intriguing opportunities to enhance the interface between operations planners and traditional OR algorithms and tools. An advanced planning system (APS) is a sophisticated software that leverages OR tools to help operations planners define, create, and interpret an operational plan. While highly beneficial, many customers are priced out of using an APS due to the ongoing costs of OR consultants for maintenance. We present SMARTAPS – a conversational system built on an agentic framework to reduce operations planners dependency on OR consultants. Our system provides operations planners with an intuitive, natural language chat interface, allowing them to make queries, perform counterfactual reasoning, receive recommendations, and execute scenario analysis (what-if and why-not analyses) on the operational plans created by the APS. SMARTAPS utilizes generic tools created by OR consultants to generalize to the specific problem. It features a module that generates and recommends new tools when faced with an unsupported query in a human-in-the-loop manner.
\end{abstract}

\section{Introduction}\label{sec:Intro}
Operations research (OR) is a discipline within applied mathematics that delivers advanced analytical tools to aid in daily decision-making and problem-solving. It has seen many successful applications and has been beneficial in different domains such as housing \cite{KILLEMSETTY2022699}, emergency response \cite{MOSTAJABDAVEH2025522, TIPPONG20221}, and healthcare operations \cite{odonnell2021operational}. Such success is largely attributed to the availability of OR tools and solvers, which have been instrumental in providing an analytical lens for decision-making.

While beneficial, using OR tools and solvers requires a high level of expertise specific to the domain and custom software. Artificial intelligence (AI) is well-positioned to support every stage of the OR project life cycle \cite{fan2024artificialintelligenceoperationsresearch}. Specifically, large language models (LLMs) have a crucial role in making these OR tools and solvers more accessible and user-friendly. For instance, propose an LLM-based framework \cite{ramamonjison-etal-2022-augmenting} has been proposed that takes in a natural language problem description and formulates the corresponding optimization model code. GPT-3.5-TURBO \cite{brown2020languagemodelsfewshotlearners} was tested on this relatively simple modeling dataset and was able to accurately convert the natural language descriptions into modeling code \cite{ramamonjison2023nl4optcompetitionformulatingoptimization}. However, LLMs have been reported to be unable to do so reliably for real-world problems that tend to be more complex \cite{wasserkrug2024largelanguagemodelsoptimization}.

In operations management, OR methods are utilized to determine the optimal decisions over a period of time (planning horizon) to produce an organization’s goods and services \cite{reid2009operations}. This is typically outlined by an operational plan, which is a detailed roadmap satisfying the requirements and strategic goals of the organization. Advanced planning systems (APSs) were developed to help create, manage, and interpret such operational plans, and are primarily used to model decisions in operations planning and supply chain management. APSs are useful for tracking and planning business details (e.g., bills of materials, customer orders, and inventory stock) and business specifications (e.g., resource shortages, equipment maintenance). They also commonly provide visualizations (e.g., Gantt charts) and a graphical user interface (GUI) so the user can inspect and manipulate a plan. However, the high costs associated with implementing and maintaining an APS often hinder their widespread adoption. These costs stem from limited automation, need for customization, and reliance on expert consultants \cite{9266587}.

In collaboration with OR experts and supply chain planners, we introduce SMARTAPS to address these shortcomings by automating analyses and making the interface for APS more intuitive. SMARTAPS is a conversational interface that allows the user to perform advanced tasks (e.g., scenario analysis, feasibility relaxation) using natural language. SMARTAPS is made up of three main modules, each built on either decoder-based \cite{jiang2023mistral7b} or encoder-based \cite{xiao2024cpackpackedresourcesgeneral} relatively light LLMs.

Tool-augmented LLM. An application programming interface (API), or tool, can be called to provide external functionalities to the LLM. For instance, answering simple mathematical questions is challenging for LLMs because it requires creativity, reasoning, and numerical calculation \cite{liu2023improvinglargelanguagemodel}. By grounding an LLM’s response to a calculator API, an LLM may act as an intermediary between the user and the tool—often resulting in more efficient and accurate responses \cite{lu2024gearaugmentinglanguagemodels, andor2019givingbertcalculatorfinding, schick2023toolformerlanguagemodelsteach}. Inspired by this work, SMARTAPS leverages customized algorithms and tools to perform complex analyses (e.g., running an optimization solver).

\section{Background in Artificial Intelligence}\label{sec:AI}
\textbf{Advanced planning systems.} Industry leaders of APS solutions include Kinaxis’ Planning One \cite{kinaxis2024planningone}, SAP’s S/4HANA \cite{sap2024s4hana}, and Oracle’s Fusion Cloud Supply Chain \& Planning \cite{oracle2024planning}. The most significant trends in the APS market are the customization capabilities for users through microservices and AI \cite{9266587}. For instance, Microsoft Copilot \cite{microsoft2024copilot} introduces an AI copilot for querying inventory stock levels using natural language. While Microsoft Copilot \cite{microsoft2024copilot} leverages AI for Dynamics 365 Supply Chain Management, it is currently only used to query details of the system. In addition to simply understanding the current data and plan, planners can use SMARTAPS for insightful tasks like operational forecasting, scenario analysis, adding restrictions, and plan comparison.

\textbf{Retrieval-augmented generation.} Retrieval-augmented generation (RAG) is a technique that retrieves relevant information from an external knowledge base to improve an LLM’s applicability in knowledge-intensive domains \cite{gao2024retrievalaugmentedgenerationlargelanguage}. RAG has been reported to improve the correctness of an LLM’s output in knowledge-intensive tasks such as generating preoperative clinical instructions \cite{ke2024developmenttestingretrievalaugmented} and scientific literature question answering \cite{lala} specific to operations management, this technique could be used to search for available and relevant code or information to help an LLM answer questions about an APS. Real-world applications of APS require more sophisticated analyses beyond what a general external database can provide. Thus, SMARTAPS uses retrieval to select the most relevant customized tools from a catalog carefully curated by an OR consultant.

\section{Methodology}\label{sec:Method}
\subsection{System Overview}
Operations planners and supply chain experts from industry were consulted throughout the iterative design process of SMARTAPS, from building the database of useful tools to designing the UI. Based on their requirements and feedback, we developed our system using a technology stack of Chainlit (chat interface), Python, Poetry (dependency management), and ChromaDB (database).

Leveraging Chainlit on the client side, a chat interface is used to accept natural language as input to interact with an APS. The chat interface and other functionalities are powered with MISTRAL-7B-INSTRUCT-V0.1 \cite{jiang2023mistral7b}, a relatively lightweight LLM using the Hugging Face inference endpoint. We employ ChromaDB for its capability to store embedding vectors for tool retrieval. BGE-LARGE-EN-V1.5 \cite{xiao2024cpackpackedresourcesgeneral} is the embedding model deployed on a single Tesla P100 PCIe 16GB GPU and used by ChromaDB to create the embedding vectors. To solve and analyze the underlying optimization model, we utilize Huawei Cloud’s OptVerse AI Solver \cite{li2024machinelearninginsidesoptverse}.

\subsection{System Implementation}
Figure 1 presents our interactive chat interface that facilitates conversations between the user (Planner) and SMARTAPS (OptVerse AI). Users pose questions and receive answers in various formats like text, tables, or graphs. A notable feature is the detailed step-by-step procedure list, which reveals the process behind the response (Figure 1 – center expanded ``Took 4 steps" pane). Additionally, the interface includes a task list on the right panel (Figure 1), visually indicating the history and status of tasks.

\begin{figure}
    \centering
    \includegraphics[width=1\textwidth]{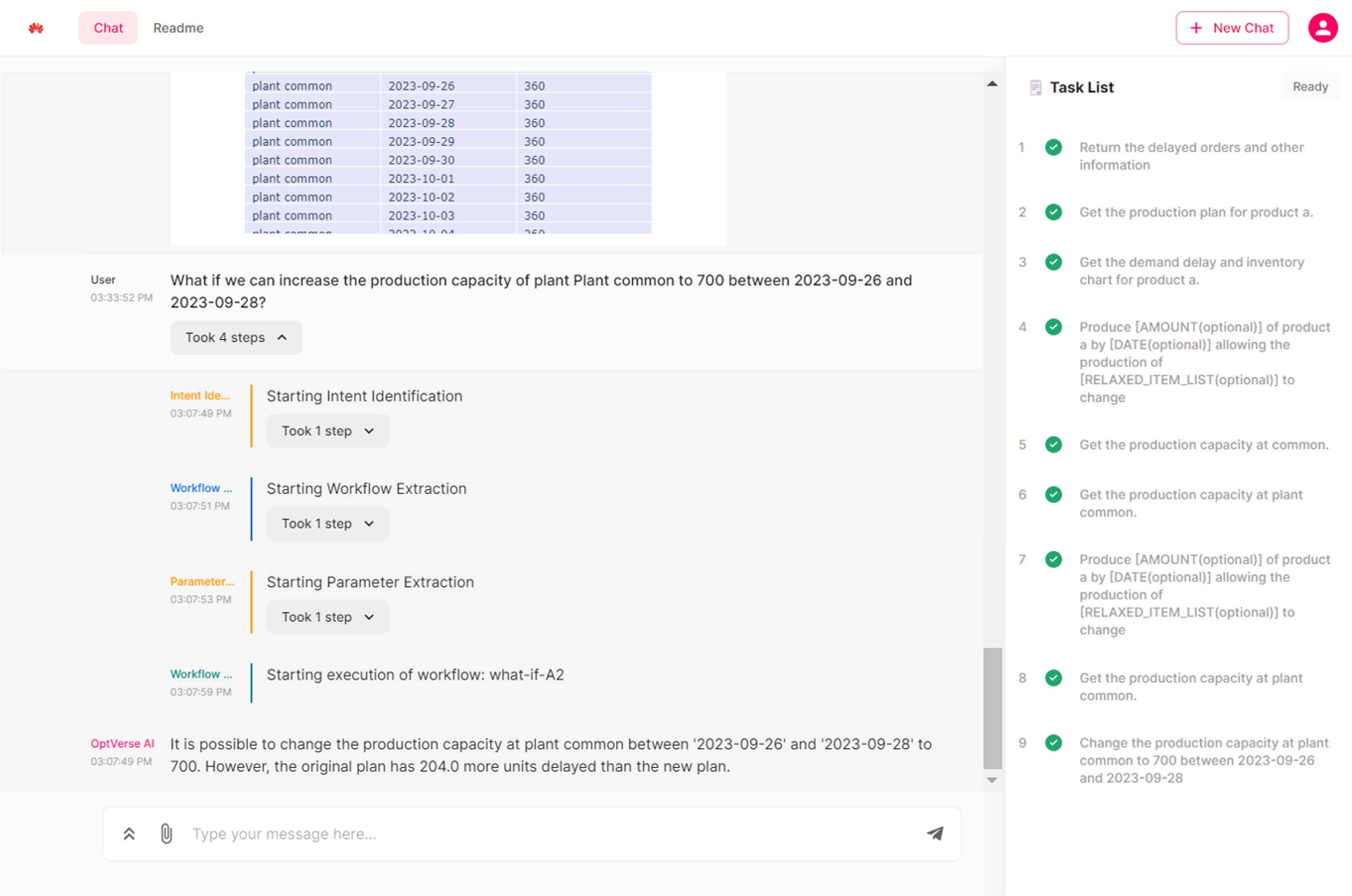}
    \caption{SMARTAPS planning application. On the left is the chat interface, where the top shows a blue table detailing the operations plan at a given plant. The expanded ``Took 4 steps" shows the process being executed. The right panel contains the task list that logs the tools executed throughout the conversation.}
    \label{fig:fig1}
\end{figure}

Figure 2 demonstrates the SMARTAPS framework at a high level. The framework allows a planner to interact with an APS through a chat interface via APIs created by the OR consultant. The framework contains three main modules: (1) conversation manager, (2) tool retriever, and (3) tool manager.

\begin{figure}
    \centering
    \includegraphics[width=1\textwidth]{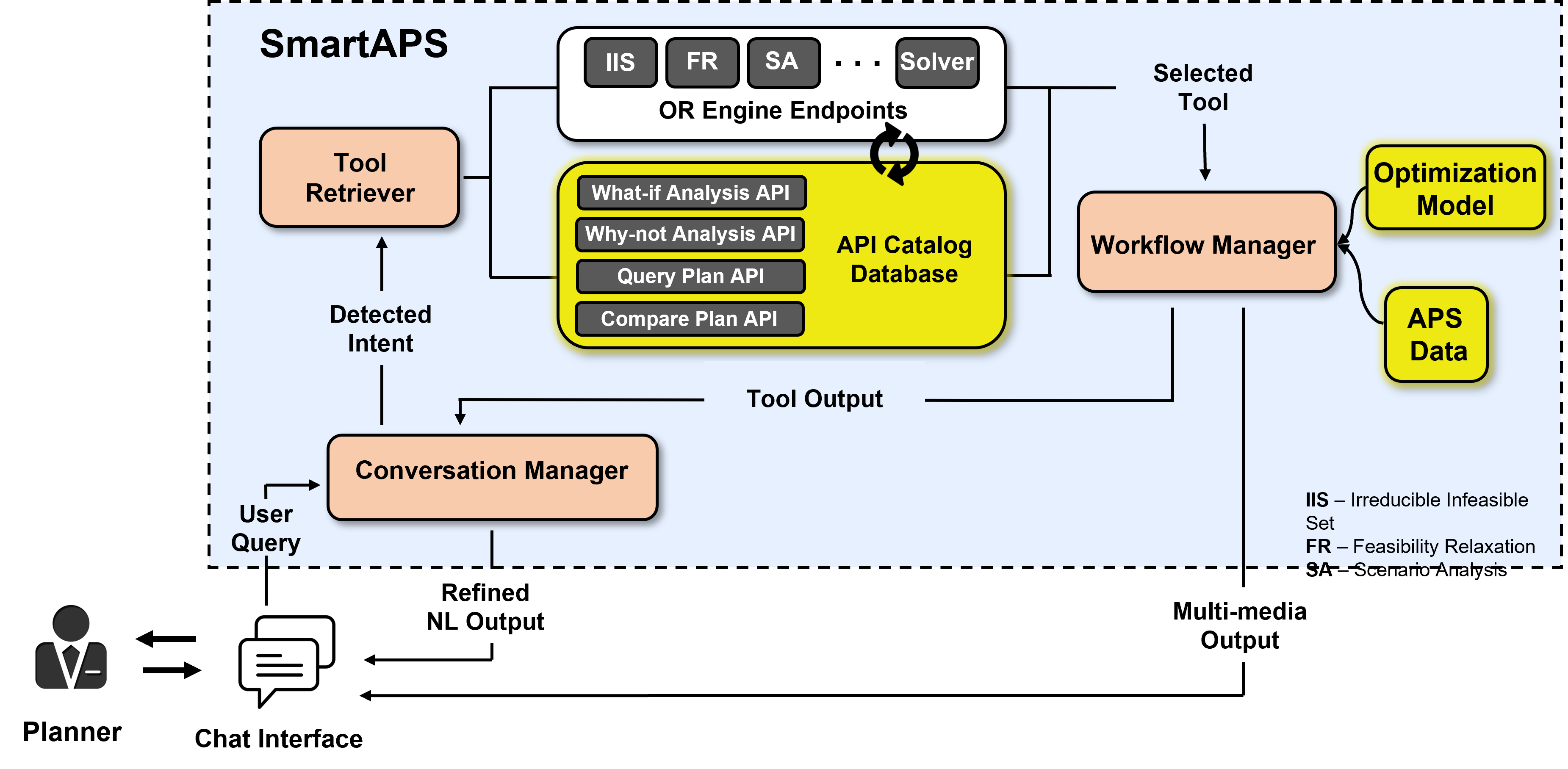}
    \caption{The SMARTAPS framework. The \textbf{conversation manager} handles the interaction with the user by detecting the user intent and refining the corresponding tool output response. The \textbf{tool retriever} selects the required tool from the API catalog database. Some tools would interact with OR engine endpoints (e.g., optimization solver) which must be deployed and accessible by the system. The \textbf{tool manager} detects the corresponding optimization model, APS data, and API input parameters from the conversation history and executes the tool.}
    \label{fig:fig2}
\end{figure}

At a high level, the conversation manager (Section 4.1) keeps track of the state of the conversation (e.g., user intent, message log) and operation (e.g., current plan, APS data, optimization models). Specifically, it uses an LLM for understanding the intent of the user and explaining the solution within the context of the conversation. The tool retriever (Section 4.2) uses an encoder to compute the sentence embedding from the natural language user query, upon which semantic textual similarity search is utilized to retrieve the most relevant tool API. The tool manager (Section 4.3) receives the retrieved API, and an LLM is used to extract the required input parameters from the user query in the context of the conversation. The tool manager will then execute the tool API and return the output to the session manager.

\textbf{Tool API.} The tools are saved as one collection in ChromaDB but may be described as five categories: query plan, why-not, what-if, compare plan, and display plan. \textit{Query plan} tools inspect the operational plan and data (e.g., How many sets of tires are made today?). \textit{Why-not} tools analyze scenarios where new requirements need changes in the optimization model (e.g., I want to only use the plant in Vancouver). \textit{What-if} tools analyze scenarios where the APS data changes (e.g., How would receiving 100 kg of natural rubber on 2024-04-17 impact my plan?). \textit{Compare plan} tools compare two generated plans (e.g., How many more tires are produced in the new plan?). \textit{Display plan} tools generate tables and charts for the user (e.g., Show me the operations plan).

Figure 3 shows the contents of a tool contract that specifies the function, requirements, and outputs of the tool. Each contract contains a description, example queries, NL output, function call, input, and output. The description and example queries are concatenated and used for tool retrieval. The ``NL output" is a template used to generate an initial structured natural language response. The ``function call" is used by the tool manager to execute the tool. The ``input" and ``output" schemas provide the user with information about the request to make and the response to expect from the tool. Notably, the output schema may return a dataframe that Chainlit can render into a plot, table, or figure.

\begin{figure}
    \centering
    \includegraphics[width=0.5\textwidth]{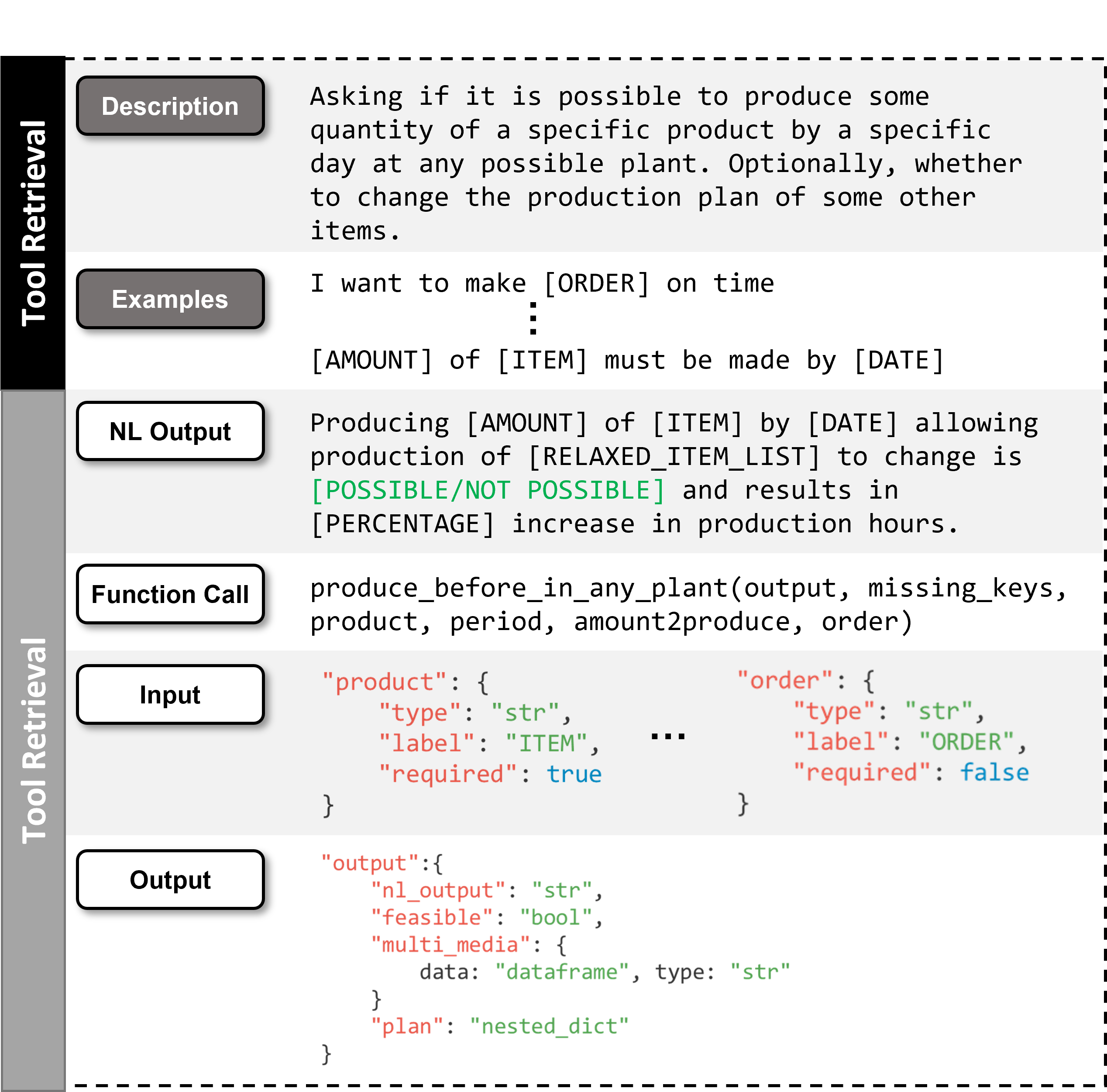}
    \caption{Example of an API contract that provides the description, examples, natural language output, function call, input schema, and output object pf the API.}
    \label{fig:fig3}
\end{figure}

\section{System Components Details}\label{sec:Details}
\subsection{Conversation Manager}
Figure 4 demonstrates the two tasks of the conversation manager: (1) user intent detection (top), and (2) refine the tool output (bottom) and return a conversation-grounded response. Both tasks are executed by prompting an LLM (in our case, we use MISTRAL-7B-INSTRUCT-V0.1 \cite{jiang2023mistral7b}) with the prompts as described in Figure 4 to the right of their corresponding tasks.

Intent detection aims to categorize the user query into one of two INTENT\_OPTIONS, either categorized as CASUAL\_CONVERSATION or OPERATIONS\_PLANNING, where CASUAL\_CONVERSATION is to continue the current conversation without the need to execute a new tool. OPERATIONS\_PLANNING is when the user has a request about the APS. The prompt to extract this intent is shown in the green classification prompt of Figure 4.

Response refining. If the intent is to continue a casual conversation, the LLM will reference the conversation when responding to the user query. Conversely, if the intent is to ask something about operations planning, the tool retriever will be called. The output object returned to the conversation manager may contain some graphics (e.g., table), which can be rendered using Chainlit. Each output object also contains a natural language response that must be refined and returned to the user. These natural language outputs are simple (e.g., ``5 hours extra delay"). The tool API does not have knowledge of the state of the conversation. Therefore, to refine the response to be suitable within the context of the conversation, the tool output and the conversation log are passed to the LLM as outlined in the blue refine response prompt of Figure 4. For example, the tool output ``5 hours extra delay" may be refined to be ``The expected delivery time for Order A will be delayed by 5 hours if Order B is prioritized."

\begin{figure}
    \centering
    \includegraphics[width=1\textwidth]{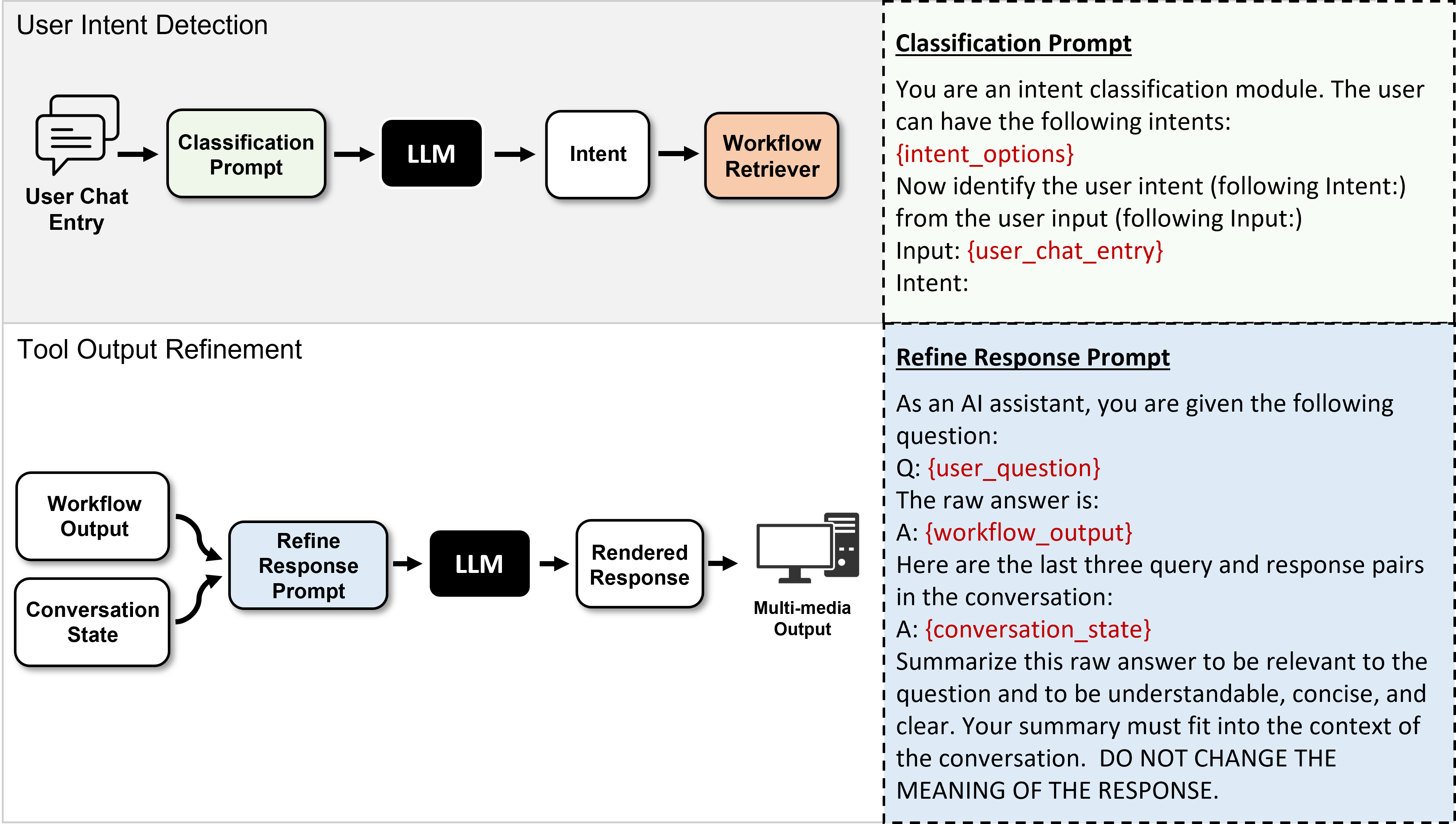}
    \caption{\textbf{Conversation Manager.} It is responsible for user intent detection (top) and tool output refinement (bottom). Both tasks leverage LLMs - the prompt template is shown to the right of each corresponding task. When refining the tool output (bottom), the tool output is the output of calling a tool API. Thus, the entire conversation manager module can be seen as a tool-augmented LLM that calls a tool and leverages it to enhance its response.}
    \label{fig:fig4}
\end{figure}

\subsection{Tool Retriever}
Figure 5 details the components of the tool retriever. The concatenated description and examples of each tool API are encoded into low-dimensional dense vectors using BGE-LARGE-EN-V1.5 \cite{xiao2024cpackpackedresourcesgeneral} when the server is initialized. When a query is received by the tool retriever, BGE-LARGE-EN-V1.5 is used once again to convert the user query into an embedding vector. Then, semantic similarity (Squared L2 norm) is calculated between the user query embedding and the embeddings representing each tool API in the ChromaDB collection, and the most similar tool is selected (minimum distance). Table 1 presents the performance of this retrieval method on an annotated test set containing 150 instances of user queries for APIs created for our case study.

\begin{figure}
    \centering
    \includegraphics[width=0.6\textwidth]{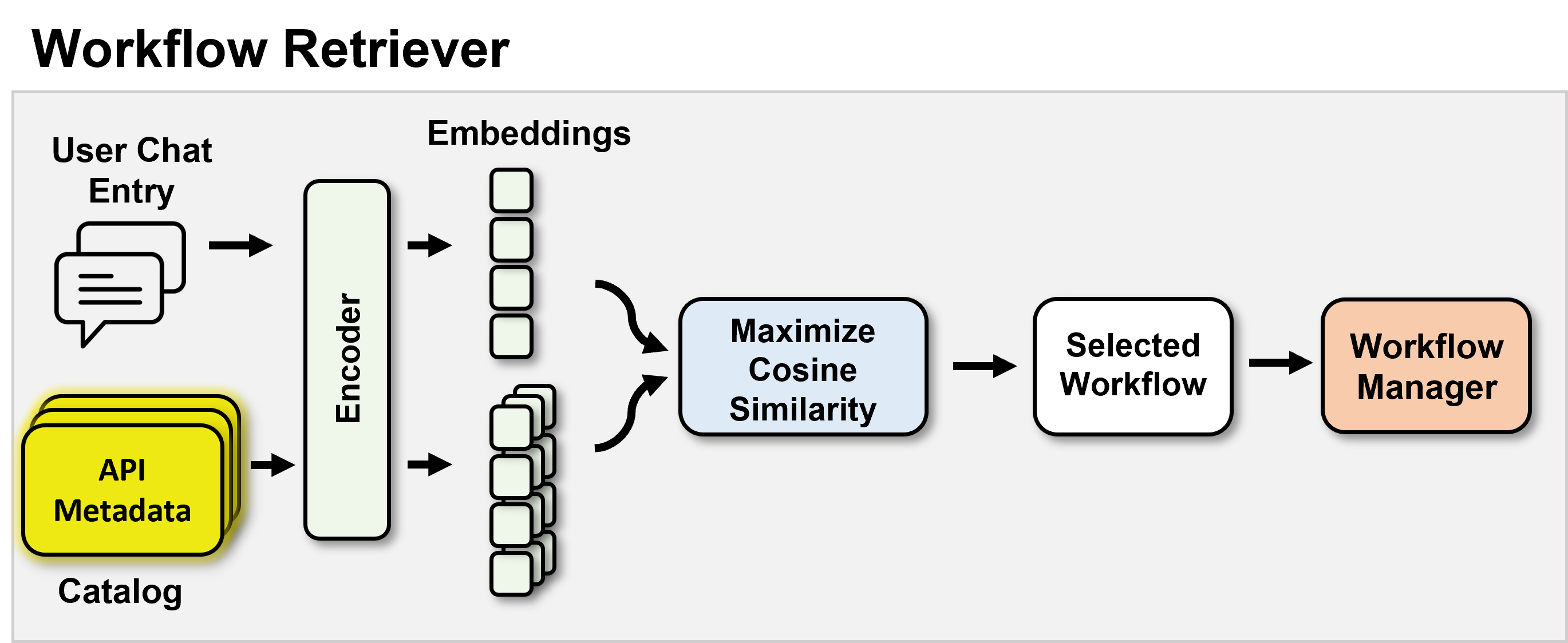}
    \caption{\textbf{Tool Retriever.} Retrieves the tool most relevant to the input query. Semantic similarity is calculated between the query and catalog of API metadata.}
    \label{fig:fig5}
\end{figure}

\subsection{Tool Manager}
Figure 6 shows the details of the tool manager. Each tool has an API contract that defines the tool API. Each contract details the accepted input parameters. Any required input parameters must be extracted either from the user query or inferred from the conversation, model, or APS data. If the tool manager is unable to infer a required missing parameter, it would return the missing parameter to the conversation manager. The conversation manager will ask clarifying questions to confirm that the retrieved tool is correct and attempt to infer the missing parameter. The prompt for extracting input parameters is shown in the green parameter extraction prompt in Figure 6. The blue prompt is used to identify the model that is associated with the user query. Within a conversation, there may be multiple models and/or APS data saved. The name (or ID) of the model is detected from the conversation and used to determine the model that corresponds to the current user query. The APS data is extracted in a similar manner. After selecting the correct model, APS data, and extracting the required input parameters, the tool manager executes the selected tool and returns the output to the conversation manager.

\begin{figure}
    \centering
    \includegraphics[width=1\textwidth]{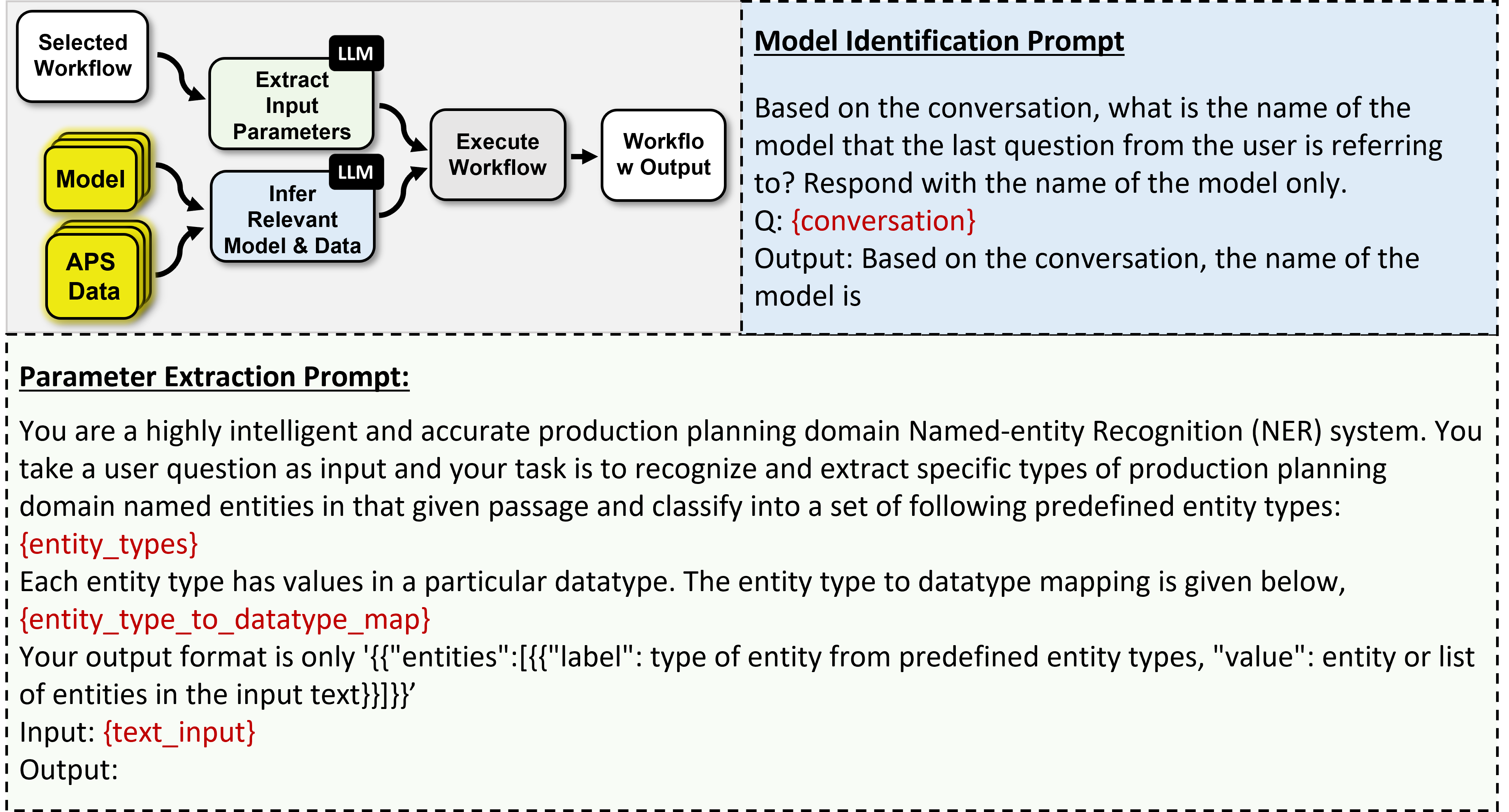}
    \caption{\textbf{Tool Manager.} It is responsible for selecting the relevant model and APS data, extracting the input parameters for the selected tool, and executing the tool. The tool manager also infers missing parameters from the user query and the conversation history.}
    \label{fig:fig6}
\end{figure}

\section{Case Study}\label{sec:Case}
Production planning, a key APS application, demands daily updates by planners to accommodate new information from manufacturing, warehouse, logistics, and sales, alongside managing requests like tight customer deadlines, machine downtime, and material sharing across plans. Addressing these requests and updating the production plan necessitates thorough analysis.

Through our discussions with Huawei’s supply chain planners and observations of their workflows, we identified their primary challenge with APS: a heavy reliance on OR consultants to conduct analyses. The most common types of analyses required by planners include finding reasons for customer order production delays and identifying resolutions.

To answer production planners' needs with SMARTAPS, we asked OR consultants to develop tools with APIs and API contracts specifically designed for production planning. The tool categories and the number of instances for each category are presented in Table 1.

We conducted a user study by deploying SMARTAPS in a realistic production planning scenario and granting planners access to it. They reported that SMARTAPS enabled them to query plans more efficiently and more readily identify the reasons for customer order production delays. Users particularly highlighted the advantages of supporting 'why-not' and 'what-if' analyses, which could reduce the time required for analysis from potentially 1-2 days—due to dependence on OR consultants—to just a few hours. Overall, the feedback from both production planners and OR consultants was positive, and we are collaborating with them to integrate SMARTAPS into their daily tool stack.

\subsection{Limitation and Future Works}
Some limitations of SMARTAPS highlight important research directions that would greatly benefit systems built upon a tool-augmented LLM like SMARTAPS. Tool-augmented frameworks must have APIs to call; they are unable to respond to requests not covered by the APIs. Especially in a technical domain like OR, APIs must be created and customized by experts. Some sophisticated code generation methods should be investigated for their ability to automatically create these advanced APIs with complex algorithms.

Optimization solvers often take a long time when calculating the optimal solution. In practice, due to the solve time, especially for larger operations, many support consultants submit a job overnight to the APS that requires a new solution to be calculated. The plan is only ready to be reviewed by the next morning. A task manager can be incorporated into SMARTAPS to keep track and allow jobs to be run in parallel.

Finally, the conversation is currently between the system and one user. In real-world operations, it is common to have multiple planners, each with different objectives. Multi-user approaches should be explored to handle this scenario.

\section{Conclusion}\label{sec:Conclusion}
In this paper, we present a demonstration of SMARTAPS, a useful and intuitive chat interface that helps users interact with an APS using natural language. It is built upon a tool-augmented LLM, and we present three main modules that have been developed using learning-based NLP methods, inspired by relevant prior works like retrieval-augmented generation and tool-augmented LLM. We will continue iteratively improving SMARTAPS based on feedback from real-world planners, with the goal of ensuring that it helps planners more efficiently manage their operations, thereby reducing OR consultation costs and turnaround time.

\bibliographystyle{plain}
\bibliography{ref}
\end{document}